\documentclass[sigconf]{aamas}  
\usepackage{balance}  

\renewcommand\footnotetextcopyrightpermission[1]{} 
\usepackage{booktabs}
\usepackage{times}
\usepackage{xcolor}
\usepackage{soul}
\usepackage[utf8]{inputenc}
\usepackage{subcaption}
\usepackage{comment}
\usepackage{helvet}
\usepackage{courier}
\usepackage{array,booktabs,multirow}
\usepackage{amsmath,amssymb,amsthm}
\usepackage{latexsym}
\usepackage{amsfonts}
\usepackage{xspace}
\usepackage{graphics}
\usepackage{multirow}
\usepackage{graphicx}
\usepackage{todonotes}
\usepackage[all]{xy}
\usepackage{url}

\acmDOI{}  
\acmISBN{}  
\acmConference{Technical Report}{}{}  
\acmYear{}  
\copyrightyear{}  
\acmPrice{}  

\def\ProofVersion{1}

\theoremstyle{definition}

\newcommand{\acceptSemantics}{\ensuremath{\operatorname{Deg}_S}}
\newcommand{\bag}{\ensuremath{\mathcal{\textbf{A}}}}

\newcommand{\weight}{\ensuremath{\mathcal{\operatorname{w}}}}
\newcommand{\arguments}{\ensuremath{\mathcal{A}}}
\newcommand{\attacks}{\ensuremath{\mathcal{R}}}
\newcommand{\supports}{\ensuremath{\mathcal{S}}}
\newcommand{\diff}[1]{\ensuremath{\frac{\mathrm{d}#1}{\mathrm{d}t}}}
\newcommand{\diffp}[2]{\ensuremath{\frac{\mathrm{d}#1}{\mathrm{d}#2}}}

\newcommand{\energysolution}{\ensuremath{\sigma^\bag}}

\begin{document}

\title{Extending Modular Semantics for Bipolar Weighted Argumentation
\ifnum\ProofVersion=1
(Technical Report)
\fi}

\author{Nico Potyka}
\affiliation{%
  \institution{University of Osnabr\"{u}ck, Institute of Cognitive Science, Germany}
}
\email{npotyka@uos.de}

\begin{abstract}
Weighted bipolar argumentation frameworks offer a tool for
 decision support and social media analysis.
Arguments are evaluated by an iterative procedure that takes
initial weights and attack and support relations into account. 
Until recently, convergence of these iterative procedures was not very
well understood in cyclic graphs.
Mossakowski and Neuhaus recently introduced a unification of different approaches
and proved first convergence and divergence results.
We build up on this work, simplify and generalize convergence results and 
complement them with runtime guarantees.
As it turns out, there is a tradeoff between semantics' convergence guarantees
and their ability to move strength values away from the initial weights. 
We demonstrate that divergence problems can be avoided without this tradeoff
by continuizing semantics. 
Semantically, we extend the framework with a Duality property that assures 
a symmetric impact of attack and support relations.
We also present a Java implementation of modular semantics and explain
 the practical usefulness of the theoretical ideas. 
\end{abstract}

\maketitle

\section{Introduction}

\emph{Abstract argumentation} \cite{dung1995acceptability} allows modeling arguments
and their relationships in order to decide which arguments should be accepted and which should be rejected.
We focus on weighted bipolar argumentation frameworks here that start with an
initial weight of arguments and adapt this weight based on the strength of their attackers and
supporters \cite{baroni2015automatic,rago2016discontinuity,amgoud2017evaluation,mossakowski2018modular,potyka2018Kr}.
These frameworks can be applied to tasks like decision support \cite{baroni2013argumentation,rago2016discontinuity}, social media analysis  \cite{leite2011social,alsinet2017weighted} or information retrieval
\cite{thiel2017web}.
Initial weights can be defined manually based on the reputation of arguments' sources or
computed automatically based on statistics like
the success rate of a source (in decision support) or the number of likes or retweets of an argument
(in social media analysis).
Sentiment analysis tools can be used to extract attack and support relations automatically as well
\cite{alsinet2017weighted}. 

Mossakowski and Neuhaus recently introduced
a unification of different approaches by decomposing
their semantics into an aggregation function that aggregates
the strength of attackers and supporters and an influence
function that adapts the initial weight based on the aggregate
\cite{mossakowski2018modular}.
Different combinations of aggregation and influence functions yield different semantics
from the literature and axioms proposed in 
\cite{amgoud2016axiomatic,amgoud2016evaluation,amgoud2017evaluation}
can be related to elementary properties of these functions.
\cite{mossakowski2018modular} also proved first results
about the convergence of bipolar weighted argumentation models in cyclic graphs.
Note that convergence is essential to obtain final strength values here.
\cite{mossakowski2018modular} gave convergence results for sum- and max-based aggregation functions and 
influence functions whose derivatives can be bounded. 

We will show that all
these results can be seen as special cases of the Contraction Principle
from Real Analysis \cite{rudin1976} and can be generalized in a uniform way 
by replacing the assumption of bounded derivatives from \cite{mossakowski2018modular} with Lipschitz 
continuity. This allows generalizing the convergence results and to add runtime guarantees. 
However, we also show that convergence guarantees derived from the contraction principle are bought at the expense of open-mindedness.
That is, as the convergence guarantees of a semantics obtained from the contraction principle get stronger, its ability to change the initial weights gets weaker. We also give some new 
divergence examples based on a family of graphs from \cite{mossakowski2018modular}.
In order to avoid the tradeoff between convergence guarantees and open-mindedness of semantics,
we can continuize semantics as proposed in \cite{potyka2018Kr}. We demonstrate that the observed 
divergence problems can be solved by continuization and, thus, give some additional empirical evidence for
the robustness of continuous models.
Subsequently, we integrate the recently introduced Duality property \cite{potyka2018Kr} into the 
framework by Mossakowski and Neuhaus by relating it to elementary properties
of the aggregation and influence function. Finally, we present an implementation of Modular semantics
in the Java library Attractor\footnote{\url{https://sourceforge.net/projects/attractorproject}} \cite{potykatutorial} and illustrate the practical usefulness of modular semantics. 

\section{BAGs and Modular Semantics}

We consider \emph{weighted bipolar argumentation graphs (BAGs)} as considered in \cite{amgoud2017evaluation} and \cite{mossakowski2018modular}.
\begin{definition}[BAG]
A \emph{BAG} is a tuple $\bag = (\arguments, \weight, \attacks, \supports)$, where
$\arguments$ is an $n$-dimensional vector of arguments, $\weight \in [0,1]^n$ is a weight vector that associates an initial weight $w_i$ with every argument $\arguments_i$ and $\attacks$ and $\supports$ are binary relations on $\arguments$ called \emph{attack} and
\emph{support}.
\end{definition}
The \emph{parent vector} $g_i \in \{-1,0,1\}^n$ of argument $\arguments_i$ is the vector
with entries $g_{i,j}=-1$ ($1$) iff $(\arguments_j, \arguments_i) \in \attacks$ ($(\arguments_j, \arguments_i) \in \supports$). 
We visualize BAGs by means of directed graphs, where nodes show the arguments with their initial weights, solid edges denote attacks and dashed edges denote supports.
We let $\textit{indegree}(\arguments_i) = \sum_{j=1}^n |g_{i,j}|$
be the number of attackers and supporters of $\arguments_i$. 
\begin{example}
\label{example_bag_1}
Figure \ref{example_bag_1_graph} shows the directed graph for the BAG $\big((a, b, c), (0.6, 0.9, 0.4), \{(a,b), (a,c)\}, \{(b,c), (c,b)\}\big)$.
 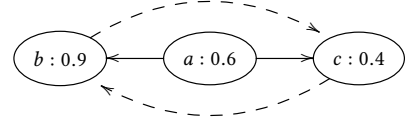
\begin{figure}[tb]
\begin{center}
		  \scalebox{.9}{
				\xymatrix{
	*+++[o][F-]{b: 0.9} \ar@{-->}@/^2pc/[rr] &
	*+++[o][F-]{a: 0.6}  \ar@{->}[l] \ar@{->}[r]   & 
	*+++[o][F-]{c: 0.4} \ar@{-->}@/^2pc/[ll]
			}	
		}
\end{center}
	\caption{Graph for Example \ref{example_bag_1}. \label{example_bag_1_graph}}
\end{figure}
The parent vector of $b$ is
$g_2 = (-1, 0, 1)$ and shows that $b$ is attacked by $a$ and supported by $c$. Hence,  $\textit{indegree}(b) = 2$.
\end{example}
Given a BAG $\bag$, we want to assign a strength value to every argument.
This can be accomplished by means of different \emph{acceptability semantics} \cite{amgoud2017evaluation}.
These semantics are usually based on an iterative update procedure that may or may not converge.
Therefore, we follow \cite{mossakowski2018modular} and regard acceptability semantics as partial functions.
\begin{definition}[Acceptability Semantics]
An acceptability semantics is a partial function $\acceptSemantics$ that maps a BAG $\bag = (\arguments, \weight, \attacks, \supports)$ with $n$ arguments
to an $n$-dimensional vector $\acceptSemantics(\bag) \in [0,1]^n$ or to $\bot$ (undefined).
If $\acceptSemantics(\bag) \neq \bot$, we call the $i$-th component $\acceptSemantics(\bag)_i$ the \emph{final strength} or \emph{acceptability degree} of $\arguments_i$.
\end{definition}
A \emph{modular acceptability semantics} as introduced in \cite{mossakowski2018modular} is an acceptability semantics that works by first aggregating the strength of attackers and supporters
and then adapting the initial weight based on the aggregated value.
This is accomplished by aggregation and influence functions, which satisfy some additional properties that guarantee that axioms from \cite{amgoud2017evaluation} are satisfied.
Even though all axioms are interesting semantically, we will restrict to a subset here in order to keep the presentation simple and more general.

The aggregation and influence functions in \cite{mossakowski2018modular} were supposed to be continuous.
We make a stronger assumption here and assume that they are \emph{Lipschitz-continuous}.
Intuitively, this means that the growth of these functions is bounded by a constant.
Lipschitz-continuity
is also implied by the convergence conditions (bounded derivatives) in \cite{mossakowski2018modular},
so we do not restrict the generality of our convergence investigation.
Formally, a function $f: X \rightarrow Y$ is called Lipschitz-continuous with
Lipschitz constant $\lambda$ iff $\|f(x) - f(y)\|_Y \leq \lambda \|x - y\|_X$. 
The sets $X$ and $Y$ will contain real numbers, vectors or matrices here.
We consider the maximum norm for matrices
defined by  $\| A \| = \max \{\sum_{j=1}^m |a_{i,j}| \mid 1 \leq i \leq n\}$
for an $m\times n$-matrix $A = (a_{i,j})$.
That is, $\| A \|$ is the largest absolute row sum in $A$.
For the special case that $x \in \mathbb{R}^n$ is a vector (an $n\times 1$-matrix), 
$\| x \|$ is the largest absolute value in $x$.
Notice that using the maximum norm does not mean any loss of generality
 because all norms are equivalent in $\mathbb{R}^n$ \cite{rudin1976} (the difference between two norms
can be bounded by a constant factor).

The aggregation function requires information about the attackers and supporters, the influence function requires information about the initial weight.
We regard this information as parameters of the function.
We also have to express that the aggregation function depends only on the parents.
As discussed in \cite{mossakowski2018modular}, this demand corresponds to the directionality axiom from 
\cite{amgoud2017evaluation}.
In order to phrase directionality, we define an equivalence relation for every parent vector $v \in \{-1, 0, 1\}^n$.
Two (strength) vectors $s_1, s_2$ are called equivalent with respect to a parent vector $v$, written as $s_1 \equiv_v s_2$ iff
$s_{1,i} = s_{2,i}$ whenever $v_i \neq 0$. That is, only the strength values of parents matter, all 
other strength values are ignored.

In the following, for a function $f$, we let $f^k$ denote the function that is obtained by applying $f$ $k$ times, that is, $f^1 = f$ and $f^{k+1} = f^{k} \circ f$.
Applying our update function repeatedly to the initial weights yields a sequence of strength vectors.
The final strength values are defined as the limit of this sequence if it exists.
Thus, convergence guarantees of update functions correspond to completeness guarantees of semantics.
As usual, we say that an n-dimensional sequence $(s_n)_{n \in \mathbb{N}}$, $s_n \in \mathbb{R}^n$, converges to $s$,
denoted as $\lim_{n \rightarrow \infty} s_n = s$,
iff the real sequence $(\| s_n - s \|)_{n \in \mathbb{N}}$ converges to $0$. That is, for every $\epsilon > 0$,
there is a $N \in \mathbb{N}$ such that $\| s_m - s \| < \epsilon$ for all $m > N$.
Intuitively, this means that the $i$-th component of $(s_n)$ converges to the $i$-th component of $s$. 

We are now ready to define basic modular semantics.
\begin{definition}[Basic Modular Semantics]
\label{def_acceptability_semantics}
A semantics $\acceptSemantics$ is called a basic modular semantics if there exists 
\begin{enumerate}
	\item an aggregation function $\alpha_v: [0,1]^n \rightarrow \mathbb{R}$ such that for all 
	\emph{parent parameters} $v \in \{-1, 0, 1\}^n$ and $s, s_1, s_2 \in [0,1]^n$ 
		\begin{itemize}
			\item $\alpha_v(s_1)=\alpha_v(s_2)$ whenever $s_1 \equiv_v s_2$, \hfill  (Directionality)
			\item $\alpha_v$ is Lipschitz-continuous, \hfill(Lipschitz-$\alpha$)
			\item $\alpha_v(s)=0$ whenever $v=0$, \hfill  (Stability-$\alpha$)
		\end{itemize} 
	\item an influence function $\iota_w: \mathbb{R} \rightarrow [0,1]$ such that for all \emph{weight parameters} $w \in \mathbb{R}$
	\begin{itemize}
			\item $\iota_w$ is Lipschitz-continuous, \hfill(Lipschitz-$\iota$)
		\item $\iota_w(0) = w$ \hfill  (Stability-$\iota$)
	\end{itemize}
\end{enumerate}
and for all BAGs $\bag = (\arguments, \weight, \attacks, \supports)$, we have
\begin{align*}
\acceptSemantics(\bag) = \lim_{k \rightarrow \infty} f_S^k(\weight).
\end{align*}
where the $i$-th component of $f_S: [0,1]^n \rightarrow [0,1]^n$ is defined by
$\iota_{\weight_i} \circ \alpha_{g_i}$ for $i=1,\dots,n$.
$f_S$ is called the \emph{update function} of $\acceptSemantics$.
\end{definition}
In practice, for the $i$-th argument $\arguments_i$, its parent vector $g_i$ serves as the parent parameter of $\alpha_v$
and its initial weight $\weight_i$ serves as the weight parameter for $\iota_w$.
Stability-$\alpha$ and Stability-$\iota$ assure that the final strength of an argument without parents
will just be its initial weight. This corresponds to the stability axiom from \cite{amgoud2017evaluation}. 

Intuitively, modular semantics compute strength values iteratively.
They start with the initial strength vector $s^{(0)} = \weight$. Then, in the $k$-th step,
the strength of argument $i$ is computed by first applying the aggregation function to $s^{(k-1)}$
and then applying the influence function to $\alpha_{g_i}(s^{(k-1)})$.
That is, $s_i^{(k)} = \iota_{\weight_i}(\alpha_{g_i}(s^{(k-1)}))$ for $k > 0$.

\def\arraystretch{1.3}
\begin{table*}
	\begin{tabular}{llll}
		\hline
		\multicolumn{2}{l}{Aggregation Functions} \\[0.0cm]
		\hline
		Sum & $\alpha^\Sigma_{v}: [0,1]^n \rightarrow \mathbb{R}$ 
		& $\alpha^\Sigma_{v}(s) = \sum_{i=1}^n v_i \cdot s_i$ 
		& $\lambda_v^\Sigma = \sum_{i=1}^n |v_{i}|$ \\[0.0cm]
		Product & $\alpha^\Pi_{v}: [0,1]^n \rightarrow [-1,1]$ &
		  $\alpha^\Pi_{v}(s) = \prod_{i: v_i = -1} (1 - s_i) - \prod_{i: v_i = 1} (1 - s_i)$ &
			$\lambda_v^\Pi = \sum_{i=1}^n |v_{i}|$\\[0.0cm]
		Top & $\alpha^\textit{max}_{v}: [0,1]^n \rightarrow [-1,1]$ &
		$\alpha^\textit{max}_{v}(s) = M_{v}(s) - M_{-v}(s)$, &
		$\lambda_v^\textit{max} = \min \{2, \sum_{i=1}^n |v_{i}|\}$ \\[0.0cm]
		&& where
		$M_{v}(s) = \max \{0, v_1 \cdot s_1, \dots, v_n \cdot s_n\}$, \\[0.0cm]
		\hline
		\multicolumn{2}{l}{Influence Functions} \\[0.0cm]
		\hline
		Linear($\kappa$) & $\iota^l_w: [-\kappa,\kappa] \rightarrow [0,1]$ &
		$\iota^l_w(s) = \weight - \frac{\weight}{\kappa} \cdot \max \{0, -s\} + \frac{1-\weight}{\kappa} \cdot \max \{0, s\}$ &
		$\lambda^l_w = \frac{1}{\kappa}\max \{w, 1-w\}$\\[0.0cm]
		Euler-based & $\iota^e_w: \mathbb{R} \rightarrow [\weight^2,1]$ &
		$\iota^e_w(s) = 1 - \frac{1 - w^2}{1 + w \cdot e^s}$ &
		$\lambda^e_w = \frac{1}{4}$ \\[0.0cm]
		p-Max($\kappa$) & $\iota^p_w: \mathbb{R} \rightarrow [0,1]$&
		$\iota_\weight^p = \weight - \weight \cdot h(-\frac{s}{\kappa}) + \weight \cdot h(\frac{s}{\kappa})$ &
		$\lambda^p_w = \frac{p}{\kappa}\max \{w, 1-w\}$ \\[0.0cm]
		for $p \in \mathbb{N}$&&where $h(x) = \frac{\max\{0, x\}^p}{1 + \max\{0, x\}^p}$ 
	\end{tabular}
	\caption{Some aggregation and influence functions with corresponding Lipschitz constants.\label{fig:aggregation_influence_examples}}	
\end{table*}   
Table \ref{fig:aggregation_influence_examples}
shows some examples of different aggregation and influence functions that can be found in the literature.
\begin{proposition}
\label{lemma_derivative_bounds}
The functions in Table \ref{fig:aggregation_influence_examples} are aggregation and influence functions
as defined in Definition \ref{def_acceptability_semantics}.
In particular, they are Lipschitz-continuous with the provided Lipschitz constants.
\end{proposition}
\ifnum\ProofVersion=1
\begin{proof}
Stability and Directionality can be easily checked from the definitions.

For Lipschitz-continuity,
we will repeatedly use the fact that if the derivative of a function is bounded by $B$,
then it is Lipschitz-continuous with Lipschitz constant $B$. This can be seen
from the intermediate value theorem \cite{rudin1976}.
We will also use the fact that the derivative of a continuously differentiable
function corresponds to a matrix of partial derivatives (the Jacobian matrix) \cite{rudin1976}.

Sum: The sum-aggregation function is continuously differentiable and 
$\diffp{\alpha_{v}}{s_i} = v_{i}$.
Hence, $\|\alpha'_{v}(s)\| = \sum_{i=1}^n |v_{i}|$.

Product: The product-aggregation function is continuously differentiable and 
$\diffp{\alpha_{v}}{s_i} = -\prod_{j: v_j = -1, j \neq i} (1 - s_j)$ if $v_{i} = -1$,
$\diffp{\alpha_{v}}{s_i} = \prod_{j: v_j = 1, j \neq i} (1 - s_j)$ if $v_{i} = 1$
and $\diffp{\alpha'_{v}}{s_i}=0$ otherwise. All derivatives are bounded from above by $1$ and 
 non-zero only if $v_{i} \in \{-1, 1\}$.
Therefore, $\|\alpha'_{v}(s)\| \leq \sum_{i=1}^n |v_{i}|$.

Top: For vectors $s, s' \in [0,1]^n$, we have
$
|M_{v}(s) - M_{v}(s')|
=|\max \{0, v_1 \cdot s_1, \dots, v_n \cdot s_n\} - \max \{0, v_1 \cdot s'_1, \dots, v_n \cdot s'_n\}|
\leq |1 - 0| = 1.
$
Hence
$
|\alpha^\textit{max}_{v}(s_1) - \alpha^\textit{max}_{v}(s_2)|
= |M_{v}(s) - M_{-v}(s) - \big(M_{v}(s') - M_{-v}(s')\big)|
\leq |M_{v}(s) - M_{v}(s')| - |M_{-v}(s) - M_{-v}(s')\big)|
\leq 2.
$
If $v$ contains only $1$ ($0$) non-zero element, only one (zero)
differences can be non-zero. Therefore, the slope is bounded by $\min \{2, \sum_{i=1}^n |v_{i}|\}$.

Linear($\kappa$): the function is not differentiable at $0$.
However, the right derivative is $\frac{1-w}{\kappa}$ and the left derivative is $-\frac{w}{\kappa}$. 
Overall, the slope is bounded at every point by $\frac{1}{\kappa}\max \{w, 1-w\}$.

Euler-based: \cite{mossakowski2018modular} showed in the proof of Theorem 8
that the derivative of the Euler-based semantics is bounded strictly from above by $\frac{1}{4}$.

p-Max($\kappa$): For $p=1$, $\max\{0, x\}$ is not differentiable at $0$, 
but the slope is bounded by $p=1$ for all $x$.
For $p>1$, $\max\{0, x\}^p$ is differentiable with 
derivative $\max\{0, p \cdot x^{p-1}\}$.
Hence, the quotient rule of differentiation implies
that the derivative of $h(x)$ is 
\begin{align*}
&\frac{\max\{0, p \cdot x^{p-1}\} \cdot (1 + \max\{0, x\}^p) - \max\{0, x\}^p \cdot \max\{0, p \cdot x^{p-1}\}}{(1 + \max\{0, x\}^p)^2} \\
&\leq \frac{\max\{0, p \cdot x^{p-1}\} }{1 + \max\{0, x\}^p} -
\frac{\max\{0, x\}^p }{1 + \max\{0, x\}^p} \cdot
\frac{\max\{0, p \cdot x^{p-1}\}}{1 + \max\{0, x\}^p} \\
&\leq
p - 0
= 
p.
\end{align*}
Hence, the chain rule of differentiation implies that the derivative of $h(\frac{s}{\kappa})$ is 
$\frac{p}{\kappa}$. 
Linearity of the limit implies then differentiability of $\iota_w^p$. The derivative is piecewise linear
with a discontinuity at $0$, but the slope can again be bounded.
The derivative is $0$ for $s=0$, bounded by
$|- \weight \cdot h'(-\frac{s}{\kappa})| \leq \frac{p \cdot \weight}{\kappa}$ for $s < 0$
and bounded by
$|\weight \cdot h'(\frac{s}{\kappa})| \leq \frac{p \cdot \weight}{\kappa}$ for $s>0$. 
Overall, the derivative is bounded by $\frac{p}{\kappa}\max \{w, 1-w\}$.
\end{proof} 
\fi

All aggregation functions that we consider here work by computing an aggregated attack and support
value independently and subtracting these values.
The sum-aggregation function has been used for the Euler-based semantics in \cite{amgoud2017evaluation} and for the quadratic energy model
in \cite{potyka2018Kr}.
It aggregates strength values by adding them.
The product-aggregation function is the aggregation function
of the DF-QuAD algorithm \cite{rago2016discontinuity}.
Intuitively, the aggregate for attack and support
is initially $1$ and the aggregates are decreased by multiplying with $(1-s)$ for an attacker or
supporter with strength $s$.
The top-aggregation function has been used for the top-based semantics in \cite{amgoud2016evaluation}
for support-only graphs and has been generalized to bipolar graphs in  
\cite{mossakowski2018modular}. It considers only the strongest 
attacker and supporter.

We consider three influence functions. The linear($\kappa$) influence function has a parameter $\kappa$ 
that we call its \emph{conservativeness} for reasons that will become clear later.
The function linear(1) can be seen as the influence function
of the DF-QuAD algorithm in \cite{rago2016discontinuity}. It moves the strength to $0$ or $1$ directly
proportional to the aggregated strength values. This yields easily interpretable results, 
but requires that the aggregation function yields values between $-1$ and $1$. Hence, it cannot be combined with the sum-aggregation function. More generally, linear($\kappa$) requires that 
the aggregation function yields values between $- \kappa$ and $\kappa$.
The Euler-based influence function has been used for the Euler-based semantics in \cite{amgoud2017evaluation}. It has some nice properties but causes an asymmetry between attack and support as we discuss later. The p-Max influence function avoids this asymmetry. The p-Max influence function with $p=2$ is
 used for the quadratic energy model in \cite{potyka2018Kr}. By increasing the parameter $p$, we
increase (decrease) the influence of aggregates  larger (smaller) than $1$.
We add again a parameter $\kappa$ for 
the conservativeness.

Table \ref{fig:example_semantics}
summarizes the building blocks of the 
DF-QuAD algorithm (DFQ), the Euler-based semantics (Euler) 
and the quadratic energy model (QE). We also add a conservativeness parameter to DFQ and QE.
\begin{table}
	\begin{tabular}{lll}
		\hline
		Semantics & Aggregation & Influence \\[0.0cm]
		\hline
		DFQ($\kappa$) & Product & Linear($\kappa$)\\[0.0cm]
		Euler & Sum & Euler-based\\[0.0cm]
		QE($\kappa$) & Sum & 2-Max($\kappa$) 
	\end{tabular}
	\caption{Example semantics from the literature.\label{fig:example_semantics}}	
\end{table}

\section{Convergence and Open-Mindedness}

As shown in \cite{mossakowski2018modular}, modular acceptability semantics always converge for acyclic graphs.
The claim remains true for basic modular semantics. In fact, the limit can be computed
in linear time by a single pass trough the graph
as we explain in the following proposition. 
\begin{proposition}[Convergence and Complexity for Acyclic BAGs]
\label{prop_acyclic_iteration_scheme}
Let $\acceptSemantics$ be a basic modular semantics.
For every acyclic BAG $\bag = (\arguments, \weight, \attacks, \supports)$ with $n$ arguments, the limit
\begin{align*}
\acceptSemantics(\bag) = \lim_{k \rightarrow \infty} f_S^k(\weight).
\end{align*}
exists and can be computed by the following algorithm:
\begin{enumerate}
  \item Compute a topological ordering of the arguments
	 and set $s^{(0)} \leftarrow \weight$ and $k \leftarrow 1$.
	\item Pick the next argument $\arguments_i$ in the order and set 
	$$\acceptSemantics(\bag)_i = \iota_{\weight_i}(\alpha_{g_i}(s^{(k-1)})).$$
	\item Set $k \leftarrow k+1$ and repeat step 2 until $k>n$.
\end{enumerate}
Provided that $\alpha_g$ and $\iota_w$ can be computed in linear time, 
the algorithm runs in linear time.
\end{proposition}
\ifnum\ProofVersion=1
\begin{proof}
For the convergence proof,
we can assume w.l.o.g. that the arguments are topologically ordered
because $\bag$ is acyclic.
That is, for every edge $(\arguments_i,\arguments_j)$ in the graph (attack or support), we have $i < j$. 
We show by induction that the strength of $\arguments_i$ remains unchanged after iteration $i$.
Since $\arguments_1$ has no predecessors,  $s^{(k)}_1 = \weight$ for all iterations $k$
by stability and directionality.
Assume that the claim is true for the first $k-1$ arguments.
Then, $s^{(m)}_i = s^{(k)}_i$ for $i=1, \dots, k-1$ and all $m>k$.
That is, $s^{(m)} \equiv_{g_k} s^{(k)}$, so that directionality implies
$s^{(m)}_k = s^{(k)}_k = (\iota_{\weight_1} \circ \alpha_{g_i})(s^{(k-1)}_k)$ for all all $m>k$.

Hence, after $n$ iterations, the procedure is guaranteed to have converged.
For the runtime analysis, we can no longer assume that the arguments are topologically ordered.
However, a topological ordering can be computed in linear time \cite{cormen2009introduction}.
The naive computation of the  strength values takes quadratic time.
However, it is actually not necessary to compute the  strength for all arguments in every iteration because 
the strength of $\arguments_i$ depends only on the strength of $\arguments_1, \dots, \arguments_{i-1}$.
Hence, it suffices to compute only $s_i$ in iteration $i$.
Then the overall runtime is linear.
\end{proof} 
\fi
We will now apply the contraction principle to unify and to generalize 
the convergence guarantees from \cite{mossakowski2018modular}. 
A contraction is a Lipschitz-continuous function with Lipschitz-constant strictly smaller than $1$.
The contraction principle states intuitively
that every contraction has a unique fixed-point that can be reached by applying the function repeatedly
starting from an arbitrary point.
\begin{lemma}[Contraction Principle]
If $S$ is a complete metric space and if $f: S \rightarrow S$ is a contraction, 
then there exists one and only one $x^* \in S$ such that $f(x^*) = x^*$.
In particular, $\lim_{n \rightarrow \infty} f^n(x) = x^*$ for all $x \in S$.
\end{lemma}  
A proof of the contraction principle can be found, for example, in \cite{rudin1976}.
The set $[0,1]^n$ of strength vectors with distance $d(x,y) = \|x-y\|$ defined by the maximum norm is indeed a complete metric space.
Given a BAG with $n$ arguments such that $(\iota_{\weight_i} \circ \alpha_{g_i})$ is a contraction
for all $i=1,\dots,n$, the contraction principle guarantees that the strength values converge.
As we will explain soon, the convergence results in  \cite{mossakowski2018modular} are special cases
of the following result. In particular, we can relate convergence time to the Lipschitz-constants.
\begin{proposition}[Convergence and Complexity for Contractive BAGs]
\label{prop_convergence_criterion}
Let $\bag$ be a BAG, let $\acceptSemantics$ be a basic modular semantics and
let $\lambda_{\bag, S} = \max_{1\leq i \leq n} \lambda^\alpha_{g_i} \cdot \lambda^\iota_{w_i}$.
If $\lambda_{\bag, S} < 1$, then
the update function $f_S$ of $\acceptSemantics$ is a contraction with unique fixed point
$s^* = \acceptSemantics(\bag)$. 

Furthermore, for all $\epsilon > 0$,
$\|f^k_S(\weight) - s^*\| \leq \epsilon$ for all $k > \frac{\log \epsilon}{\log \lambda_{\bag, S}}$.
\end{proposition}
\ifnum\ProofVersion=1
\begin{proof}
First note that Lipschitz-continuous functions are closed under function composition, for
if $g_1: Y \rightarrow Z$ and $g_2: X \rightarrow Y$ are Lipschitz-continuous with Lipschitz constants $\lambda_1, \lambda_2$,
then $\|g_1(g_2(x)) -+ g_1(g_2(y))\| \leq \lambda_1 \cdot \|g_2(x) - g_2(y)\| \leq \lambda_1 \cdot \lambda_2 \cdot \| x -y\|$.
That is, $g_1 \circ g_2$ is Lipschitz-continuous with Lipschitz-constant $\lambda_1 \cdot \lambda_2$. 
Hence, $\iota_{\weight_i} \circ \alpha_{g_i}$ is Lipschitz-continuous with Lipschitz constant 
$\lambda^\alpha_i \cdot \lambda^\iota_i < 1$.
That is, $f_S$ is a contraction and the claim follows from the contraction principle.

For the convergence guarantee, note that 
$\|f(\weight) - s^*\| = \|f(\weight) - f(s^*)\| \leq \lambda_{\bag, S} \|\weight - s^*\|$.
It follows by induction that $\|f^k(\weight) - s^*\| \leq \lambda_{\bag, S}^k \|\weight - s^*\|$.
Since all strength values must be in $[0,1]$, $\|\weight - s^*\| \leq 1$. Therefore,
\begin{align*}
\|f^k(\weight) - s^*\| 
&\leq \lambda_{\bag, S}^k 
= \exp\big( k \cdot \log \lambda_{\bag, S}\big)\\
&< \exp\bigg( \frac{\log \epsilon}{\log \lambda_{\bag, S}} \cdot \log \lambda_{\bag, S}\bigg)
= \epsilon.
\end{align*}
The inequality in the second line holds because $\lambda_{\bag, S} < 1$ 
implies $\lambda_{\bag, S} < 0$. Hence, $k > \frac{\log \epsilon}{\log \lambda_{\bag, S}}$
implies $k \cdot \log \lambda_{\bag, S} < \frac{\log \epsilon}{\log \lambda_{\bag, S}} \cdot \log \lambda_{\bag, S}$
and the inequality follows because the exponential function is monotonically increasing.
\end{proof} 
\fi
Note, in particular, that
the convergence bound in the last line implies 
$\|f^k_S(\weight) - s^*\| \leq 10^{-n}$ for all $k > C \cdot n$, where C is a constant that decreases with the Lipschitz constants of the aggregation and influence functions. 
In this sense, the strength values converge in linear time. 
In order to relate Proposition \ref{prop_convergence_criterion} to the convergence results in \cite{mossakowski2018modular},
we briefly repeat them here.
\begin{proposition}[Convergence Guarantees from \cite{mossakowski2018modular}]
\label{prop_convergence_mossakowski}
Consider a BAG $\bag$ and a modular semantics that uses 
\begin{enumerate}
	\item Sum for aggregation 
and an influence function whose derivative is strictly bounded by $M$.
If the indegree of every argument in $\bag$ is bounded by $\frac{1}{M}$,
then $f^n(x)$ converges.
 \item Top for aggregation 
and an influence function whose derivative is strictly bounded by $\frac{1}{2}$.
Then $f^n(x)$ converges.
\end{enumerate} 
\end{proposition}
Both results are special cases of Proposition \ref{prop_convergence_criterion}.
For the first result, we can see from Table \ref{fig:aggregation_influence_examples}
that the Lipschitz-constant of
sum-aggregation, when applied to a particular argument, corresponds to the indegree of the argument.
That is, $\lambda^\Sigma_{g_i} =\textit{indegree}(\arguments_i)$. 
Furthermore, if the derivative of a function is  $B$, it is also Lipschitz-continuous
with Lipschitz-constant $B$. Therefore, $\lambda^\iota_{w_i} < \frac{1}{M}$.
Hence, if the maximal indegree in $\bag$ is bounded by $M$, the condition of
 Proposition \ref{prop_convergence_criterion} becomes 
$\max_{1\leq i \leq n} \lambda^{\Sigma}_{g_i} \cdot \lambda^\iota_{w_i}
< \max_{1\leq i \leq n} \frac{M}{M} = 1$
and is satisfied as well.
For the second result, note from Table \ref{fig:aggregation_influence_examples} that the Lipschitz-constant
of top-aggregation can never be larger than 2. Hence, if the derivative of the influence function is bounded by $\frac{1}{2}$,
the condition of Proposition \ref{prop_convergence_criterion} is satisfied as before.

Hence, Proposition \ref{prop_convergence_criterion} unifies the results from \cite{mossakowski2018modular}.
It is also more general and can immediately be applied to other aggregation functions like Product-aggregation.
For the influence function, it is also slightly more general in the sense that bounded derivatives imply Lipschitz-continuity,
but not the other way round. In many cases, practical influence functions will only be pointwise non-differentiable
like Linear($\kappa$) or 1-Max($\kappa$). Proposition \ref{prop_convergence_criterion} still simplifies the investigation in these cases because
we do not have to make any complicated case differentiations for such points. 
Proposition \ref{prop_convergence_criterion} implies several new convergence guarantees.
We summarize some guarantees for product-aggregation in the following corollary.
\begin{corollary}
\label{prop_convergence_dfquad_qenergy}
Consider a BAG  $\bag$ with maximum indegree $D = \max_{1 \leq i \leq n} \textit{indegree}(\arguments_i)$.
When using a modular semantics with Product-aggregation, 
the strength values are guaranteed to converge 
\begin{itemize}
 \item if the Linear($\kappa$) influence function is  used and $D < \kappa$,
 \item if the Euler-based influence function is used and $D < \frac{\kappa}{4}$,
 \item if the p-Max($\kappa$) influence function is used and $D < \frac{\kappa}{p}$.
\end{itemize} 
When all weights in $\bag$ are strictly between $0$ and $1$, then $<$ can be replaced with $\leq$
for Linear($\kappa$) and p-Max($\kappa$).

When using Sum-aggregation and p-Max($\kappa$), the strength values are guaranteed to converge if 
$D < \frac{\kappa}{p}$. Again, $<$ can be replaced with $\leq$ if all weights are strictly between $0$ and $1$.
\end{corollary}
\ifnum\ProofVersion=1
\begin{proof}
We give a proof for Sum-aggregation and p-Max($\kappa$), all other proofs are analogous. 

In general, $0 \leq \weight_i \leq 1$ and therefore $\{w_i, 1-w_i\} \leq 1$.
Hence, 
$\max_{1\leq i \leq n} \lambda^\alpha_i \cdot \lambda^\iota_i
\leq \max_{1\leq i \leq n} \frac{p}{\kappa} \cdot \textit{indegree}(\arguments_i) < 1$
and convergence follows from Proposition \ref{prop_convergence_criterion}.

If $0 < \weight_i < 1$, we have $\{w_i, 1-w_i\} < 1$.
Hence, 
$\max_{1\leq i \leq n} \lambda^\alpha_i \cdot \lambda^\iota_i
< \max_{1\leq i \leq n} \frac{p}{\kappa} \cdot \textit{indegree}(\arguments_i) < 1$
and convergence follows from Proposition \ref{prop_convergence_criterion}.

\end{proof}
\fi
In order to show that these bounds cannot be improved much further, we give some tight 
examples based on a family of
BAGs from \cite{mossakowski2018modular}. 
We denote the members of the family by $\bag(k, v_a, v_b)$.
$\bag(k, v_a, v_b)$ contains $k$ nodes $a_i$ with weight $v_a$ and $k$ nodes $b_i$ with weight $v_b$.
All $a_i$ attack all $a_j$ and all $b_i$ attack all $b_j$ (including self-attacks).
Furthermore, all $a_i$ support  all $b_j$ and all $b_i$ support all $a_j$. 
Hence, the indegree of every argument in $\bag(k, v_a, v_b)$ is $2k$ ($k$ supporters and $k$ attackers). 

\begin{figure}
	\centering
		\includegraphics[width=0.46\textwidth]{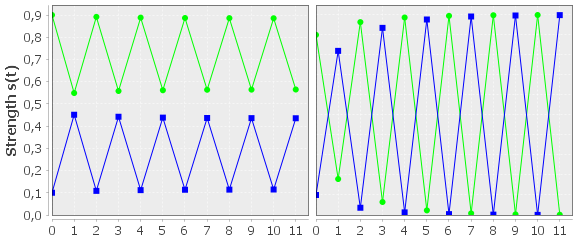}
	\caption{Divergence of QE(1) (left) and DFQ(1) (right) for $\bag(1, 0.9, 0.1)$.}
	\label{fig_divergence_example}
\end{figure}

Figure \ref{fig_divergence_example} illustrates the behaviour of DFQ(1) and QE(1) for the BAG 
$\bag(1, 0.9, 0.1)$,
where the green and blue dots show the strength of argument $a_1$ and $b_1$ over a number of iterations.
Both models start jumping between the same two states after a small number of iterations.
Since $\bag(1, 0.9, 0.1)$ has indegree $2$, this
is a tight example for DFQ(1) and QE(1) that shows that the general bounds 
given in Corollary \ref{prop_convergence_dfquad_qenergy} cannot be improved significantly. 

\begin{figure}
	\centering
		\includegraphics[width=0.46\textwidth]{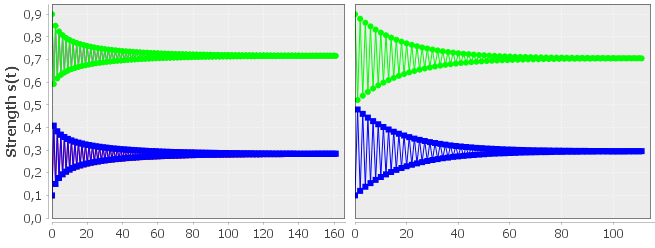}
	\caption{Convergence of QE($2.1$) (left) and DFQ($1.9$) (right) for $\bag(1, 0.9, 0.1)$.}
	\label{fig_convergence_example}
\end{figure}

As we illustrate in Figure \ref{fig_convergence_example}, we can solve the divergence problem
by increasing the conservativeness parameter $\kappa$ of the semantics.
Indeed, since increasing the conservativeness decreases the Lipschitz-constant, we can see from
Proposition \ref{prop_convergence_criterion} that the convergence guarantees improve.
However, of course, this also affects the semantics as we discuss next. 

\subsection{Open-Mindedness}

Proposition \ref{prop_convergence_mossakowski}
implies that semantics that use top for aggregation
and an influence function with derivative bounded from above strictly by $\frac{1}{2}$ 
are guaranteed to converge. Hence, when using the Euler-based influence function or influence functions 
that scale the influence of the aggregated value down by a constant $\kappa$ similar to Linear($\kappa$) and
p-Max($\kappa$), the semantics converges in general. While this is a nice guarantee, it does not 
come without cost. The bound imposed on the growth of the influence function limits the semantics' ability to adapt the initial weight as we illustrate in the following example.
\begin{example}
\label{example_open_mindedness}
Consider a BAG with one argument $a$ and $k$ arguments $b_i$ that attack $a$.
All arguments have initial weight $0.9$. Table \ref{fig:open_mindedness_example_table}
shows final strength values of argument $a$ for modular semantics with different building blocks.
Naturally, when using top for aggregation, the final strength is independent of the number of attackers.
We can also see that increasing the conservativeness parameter lets the final strength values keep
closer to the initial weights. Note also that the Euler-based semantics is extremely conservative.
\begin{table}
	\begin{tabular}{llccc}
		\hline
		 $\alpha$ & $\iota$ & $k=1$  & $k=10$  & $k=100$  \\
		\hline
		Sum & Euler & 0.862 & 0.811 & 0.811   \\
		Top & Euler & 0.862 & 0.862 & 0.862  \\
		Sum & 2-Max(1) & 0.498 & 0.012 & 0.001  \\
		Top & 2-Max(1) & 0.498 & 0.498 & 0.498  \\
		Sum & 2-Max(5) & 0.873 & 0.213 & 0.004   \\
		Top & 2-Max(5) & 0.873 & 0.873 & 0.873  \\
	\end{tabular}
	\caption{Strength values of $a$ under different semantics and increasing number of attackers $k$ for BAG from Example
	\ref{example_open_mindedness}.\label{fig:open_mindedness_example_table}}	
\end{table}
\end{example}
Arguably, a semantics should be able to move the strength values arbitrarily close
 to the extreme
values $0$ or $1$ if sufficient evidence against or for the argument is given.
We call such a semantics open-minded.
\begin{definition}[Open-Mindedness]
We say that an influence function $\iota: [l,u] \rightarrow [0,1]$ is open-minded
if $\lim_{a \rightarrow l} \iota(a) = 0$ and   $\lim_{a \rightarrow u} \iota(a) = 1$.

We call a basic modular semantics with
aggregation function $\alpha: [0,1]^n \rightarrow [l,u]$ open-minded
when its influence function restricted to the domain $[l,u]$ is open-minded.
\end{definition}
Note that we do not demand that the influence function ever yields the extreme values
$0$ or $1$ (this would be in conflict with the Resilience axiom from \cite{amgoud2017evaluation}), we only demand that it is possible to get arbitrarily close to these bounds.
For the Euler-based influence function, we have
$\lim_{a \rightarrow - \infty} \iota^e_\weight(a) = 1 - \frac{1 - \weight^2}{1 + \weight \cdot 0} = \weight^2$. Hence, the Euler-based semantics is not open-minded since it does not admit final strength values smaller
than $\weight^2$.
For example, in Table \ref{fig:open_mindedness_example_table}, the Euler-based influence function cannot yield a final strength value smaller than $0.9^2 = 0.81$.
Linear($\kappa$) and p-Max($\kappa$) are open-minded influence functions
and DFQ(1) and QE($\kappa$) are open-minded semantics. 
However, DFQ($\kappa$) is not open-minded for $\kappa>1$. 
Also, none of the semantics with general convergence guarantees 
from \cite{mossakowski2018modular} are open-minded. 
These negative results are all special cases of the following proposition.
\begin{proposition}
Consider a basic modular semantics with
aggregation function $\alpha: [0,1]^n \rightarrow [-B,B]$
and influence function $\iota$ whose Lipschitz constant is bounded by $\lambda^\iota$. 
Then for every BAG
$\bag = (\arguments, \weight, \attacks, \supports)$ with $n$ arguments, the following bound is true 
for all $i = 1,\dots, n$:
\begin{align*}
\weight_i - B \cdot \lambda^\iota \leq \acceptSemantics(\bag)_i \leq \weight_i + B \cdot \lambda^\iota.
\end{align*}
\end{proposition}
\ifnum\ProofVersion=1
\begin{proof}.
By stability-$\iota$, we have $\iota_\weight(0) = \weight$.
Hence, for all $a \in [-B, B]$, we have
$
|\iota_\weight(a) - \weight| 
= |\iota_\weight(a) - \iota_\weight(0)|
\leq  \lambda^\iota \cdot |a - 0|
\leq  \lambda^\iota \cdot B.
$ 
\end{proof}
\fi
For example, the Euler-based influence function has $\lambda^e = 0.25$. 
For aggregation with top, we have $B=1$.
Hence, when combining these two, no weight can change by more than $0.25$. 

It seems that when strong convergence guarantees can be derived from the contraction principle,
they are bought at the expense of open-mindedness. The extreme case would be the constant 
influence function $\iota_\weight(a) = \weight$ that just assigns the initial weight to every aggregate.
Its Lipschitz constant is $0$ and every basic modular semantics that uses this influence function
is guaranteed to converge trivially. 
As we let $\kappa$ in DFQ($\kappa$) and QE($\kappa$) go to infinity,
we gradually increase our convergence guarantees, but simultaneously approach the constant influence
function that leaves all weights unchanged. All currently known convergence guarantees for cyclic BAGs
seem to be 
of this kind: we buy convergence guarantees at the expense of open-mindedness.

\section{Continuous Modular Semantics}

We now look at another approach to improve convergence guarantees. Instead of making semantics more
conservative, we will adapt the update approach. Roughly speaking, we will replace coarse updates
 with more fine-grained updates. We will show that this approach leaves the semantics
unchanged in cases where we have convergence guarantees. More importantly, it can still converge to a 
fixed-point of the semantics when the original updating approach diverges. 

Roughly speaking, discrete update approaches work by applying an update
formula to the initial weights repeatedly until the process converges. In case of basic modular semantics,
the update formula is given by the function $(\iota_{\weight_i} \circ \alpha_{g_i})$.
In \cite{potyka2018Kr}, it has been proposed to use continuous models rather than discrete ones
in order to deal with cyclic BAGs. 
Continuous models can be designed in a more descriptive way than discrete models.  
To this end, the continuous change of arguments' strength based on the strength of their attackers and supporters is described by means of differential equations.
If the system of differential equations is designed carefully, it yields a unique solution
$\energysolution: \mathbb{R}^+_0 \rightarrow \mathbb{R}^n$. Intuitively,
the $i$-th component $\energysolution_i(t)$ tells us the strength of the $i$-th argument at (continuous) time $t$ 
and the final strength values correspond to the limit $\lim_{t \rightarrow \infty} \energysolution(t)$.
Just like the limit $\lim_{k \rightarrow \infty} f_S^k(\weight)$
for discrete basic modular semantics may not exist, the limit $\lim_{t \rightarrow \infty} \energysolution(t)$ may not exist. However, if we can continuize a discrete model, the discrete model can actually be
seen as a coarse approximation of the continuous model \cite{potyka2018Kr}.
In particular, the continuous model may still converge when its discrete counterpart diverges
as we will demonstrate soon. 
While there are currently no strong analytical guarantees for continuous models in cyclic BAGs, no
divergence examples have been found either and experiments
show that they can converge quickly for large cyclic BAGs with thousands of arguments.
Furthermore, sufficient conditions have been given under which discrete models can be continuized.
The results can actually be simplified and generalized to all basic modular semantics. The key property
of the aggregation and influence functions is again Lipschitz continuity.

Before stating the result, we add some explanations. 
The continuized model can be obtained as the unique solution of a system of differential equations.
The equations basically describe how the strength evolves at each current point in time based 
on the current strength. This is done by defining the derivatives of the function $\energysolution: \mathbb{R}^+_0 \rightarrow \mathbb{R}^n$. As it turns out, in order to continuize a basic modular semantics, we can just define the derivative for
the i-th strength value at time $t$
as the difference $(\iota_{\weight_i} \circ \alpha_{g_i})(\sigma(t)) - \sigma_i(t)$.
That is, as the difference between the result of applying the update function to the current state
and the state itself. Note that the difference is $0$ if
$\sigma(t)$ is a fixed-point of the function $(\iota_{\weight_i} \circ \alpha_{g_i})$. In this case, the strength value remains unchanged.
If $(\iota_{\weight_i} \circ \alpha_{g_i})(\sigma(t)) > \sigma_i(t)$, the difference, and hence the slope, will be positive and the strength value increases.
This does again make intuitively sense because the strength will be shifted towards the strength value that is desired by the update formula. 
For the case $(\iota_{\weight_i} \circ \alpha_{g_i})(\sigma(t)) < \sigma_i(t)$, the strength decreases symmetrically.
We are now ready to state the general result.
As usual, we leave out the function parameter $t$ when writing differential equations. 
\begin{proposition}[Continuizing Basic Modular Semantics]
\label{prop_continuizing_independent_semantics}
Let $\acceptSemantics$ be a basic modular semantics with aggregation function $\alpha_g$
and influence function $\iota_w$.
\begin{enumerate}
\item For all BAGs $\bag$, the system of differential equations
\begin{align}
\label{eq_continuized_odes}
\diff{\sigma_i} = (\iota_{\weight_i} \circ \alpha_{g_i})(\sigma) - \sigma_i
\end{align}
with initial conditions 
$\sigma_i(0) = \weight(i)$ for $i = 1,\dots, n$ has a unique solution $\energysolution: \mathbb{R}^+_0 \rightarrow \mathbb{R}^n$.
\item If $\energysolution$ converges and 
$s^* = \lim_{t \rightarrow \infty} \energysolution(t)$, then
$s^*$ is a fixed-point of the update function $f_S$ of $\acceptSemantics$.
\item If $\bag$ is acyclic, the discrete and continuized models converge to the same limit.
\item If $\energysolution$ converges and $f_S$ is a contraction,
then the discrete and continuized models converge to the same limit.
\end{enumerate}
\end{proposition}
\ifnum\ProofVersion=1
\begin{proof}
1. Lipschitz continuity of $(\iota_{\weight_i} \circ \alpha_{g_i})$ allows us to apply existence and uniqueness theorems for nonlinear systems of ordinary differential equations from \cite[Section 7.1.2]{polyanin2017handbook} that imply the claim.

2. If $\energysolution$ converges, then all derivatives $\diff{\sigma_i}$ must go to $0$.
  Hence, in the limit $0 = (\iota_{\weight_i} \circ \alpha_{g_i})(s^*) - s^*_i$. That is, $f_i(s^*) = (\iota_{\weight_i} \circ \alpha_{g_i})(s^*) = s^*_i$.
 
3. Analogous to the proof of Proposition 16 in \cite{potyka2018Kr},
one can show that $\energysolution$ converges to the same limit as the algorithm 
given in Proposition \ref{prop_acyclic_iteration_scheme} for acyclic BAGs.

4. If $f_S$ is a contraction, the
contraction principle implies that $f_S$ has a unique fixed-point.
Since $\energysolution$ converges to such a fixed-point according to Item 2, both models must
converge to the same limit.
\end{proof}
\fi
As opposed to the continuization result in \cite{potyka2018Kr}, the proposition does
not assume continuous differentiability of the update function
 and therefore applies to more general acceptability semantics
like the DF-QuAD algorithm from \cite{rago2016discontinuity} (DFQ(1) in Table \ref{fig:example_semantics}).
The reason that the result applies to all basic modular semantics is that they have a Lipschitz-continuous
update function, which is sufficient.

\begin{figure}
	\centering
		\includegraphics[width=0.46\textwidth]{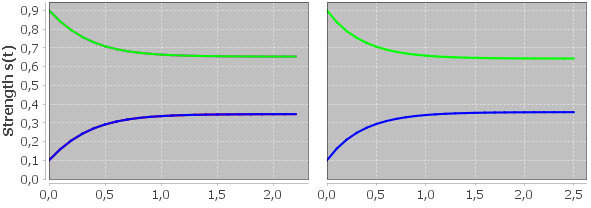}
	\caption{Convergence of Continuous QE(1) (left) and Continuous DFQ(1) (right) for $\bag(1, 0.9, 0.1)$.}
	\label{fig_continuized_convergence_example}
\end{figure}

We demonstrate in Figure \ref{fig_continuized_convergence_example} that continuizing discrete models can solve  
divergence problems. Whereas QE(1) and DFQ(1) diverged for $\bag(1, 0.9, 0.1)$ (Figure \ref{fig_divergence_example}), their continuized counterparts (Figure \ref{fig_continuized_convergence_example}) converge.
The intuitive reason for this is best explained by numerical solution techniques that approximate
the continuous model $\energysolution: \mathbb{R}^+_0 \rightarrow \mathbb{R}^n$. 
The most naive technique is Euler's method. 
In our context, it initializes the strength values with the initial conditions given by the initial 
weights. That is, $\energysolution(0) = \weight$. In order to compute $\energysolution(\delta)$ for
some small $\delta > 0$, Euler's method uses a first-order Taylor approximation. 
The first order Taylor approximation of a differentiable function $f: \mathbb{R} \rightarrow  \mathbb{R}^n$ about a point $t$ is
given as $f_i(t + \delta) \approx f_i(t) + \delta \cdot \diff{f_i}(t)$. 
Since we know $\energysolution(0) = \weight$ and $\diff{\energysolution_i}(0) = (\iota_{\weight_i} \circ \alpha_{g_i})(\weight) - \weight$,
the first-order Taylor approximation of $\energysolution_i(\delta)$ is $\weight + \delta \cdot \big( (\iota_{\weight_i} \circ \alpha_{g_i})(\weight) - \weight \big)$.  
Having obtained our approximation for $\energysolution(\delta)$,
we can move on approximating $\energysolution(2 \cdot \delta)$ analogously.
In this way, we can approximate $\energysolution(t)$ for all $t >0$ until the strength values converge.
$\delta$ is called the step-size of the approximation and we can improve the approximation quality by decreasing $\delta$.
As $\delta \rightarrow 0$, the approximation error goes to $0$ by differentiability of $\energysolution$.

\begin{figure}
	\centering
		\includegraphics[width=0.46\textwidth]{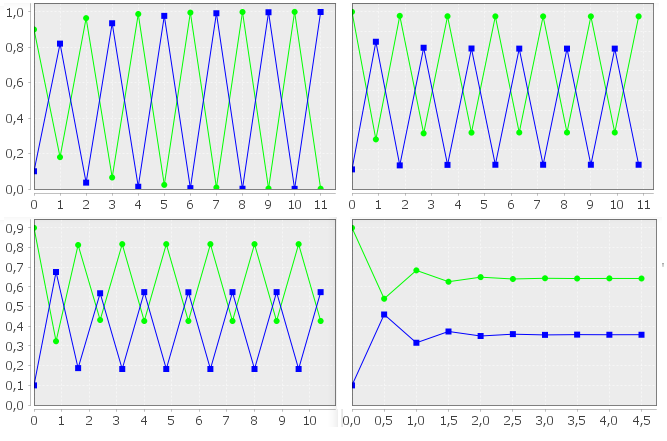}
	\caption{Approximating Continuous DFQ(1) with Euler's method for $\bag(1, 0.9, 0.1)$ with $\delta = 1$ (upper left), $\delta = 0.9$ (upper right), $\delta = 0.8$ (lower left) and $\delta = 0.5$ (lower right)
	\label{fig_continuized_approximation_example}}
\end{figure}

Interestingly, the discrete update scheme turns out to be a Taylor-approximation of the continuous model with step size $1$.
To see this, just plug in $\delta = 1$ in our formula above. Then the approximation of
$\energysolution(1)$ is $\weight + 1 \cdot \big( (\iota_{\weight_i} \circ \alpha_{g_i})(\weight) - \weight \big) 
= (\iota_{\weight_i} \circ \alpha_{g_i})(\weight)$. Notice that this is just our update formula
applied to the initial weights once. Hence, applying the update formula once can be seen as a very coarse
approximation of the continuous model at time $1$ and, more generally, applying the update formula $k$ times can be seen
as a coarse approximation of the continuous model at time $k$.
Due to this coarseness, we may actually jump from the function graph of the true solution
to the function graph of a solution for different initial conditions. 
This may cause divergence when the
algorithm starts jumping back and forth between two function graphs. 
We can avoid these jumps by decreasing $\delta$.
We illustrate this in Figure \ref{fig_continuized_approximation_example}
for DFQ(1) and the BAG $\bag(1, 0.9, 0.1)$.
As we decrease $\delta$ from $1$ to $0.8$, the oscillations already become weaker, but the step size
is not sufficiently small to avoid divergence. For $\delta = 0.5$, the oscillations die out and the 
true limit shown in Figure \ref{fig_continuized_convergence_example} is eventually reached.

Of course, we refer to Euler's method only for didactic reasons. The results in Figure \ref{fig_continuized_convergence_example} were computed using the classical Runge-Kutta method RK4 that 
provides much stronger approximation guarantees \cite{polyanin2017handbook}.

\section{Duality Property}

In order to complement the semantical properties of basic modular semantics, 
we now generalize a symmetry property introduced in 
\cite{potyka2018Kr} to the setting from \cite{mossakowski2018modular}.
Intuitively, our symmetry property should assure that attackers move
the strength from the initial weight towards $0$ in the same way as supporters move the
strength from the initial weight towards $1$. 
This can be described by constraints on the
aggregation and influence functions as follows.
\begin{definition}[Duality]
\label{def_duality}
A basic modular semantics satisfies Duality iff
\begin{enumerate}
	\item $\alpha_g(s) = -\alpha_{-g}(s)$ for all $s \in [0,1]^n$ and
	\item $1 - \iota_{(1 - \weight)}(a) = \iota_{\weight}(-a)$ for all $\weight \in [0, 1]$. 
\end{enumerate}
\end{definition}
The aggregation condition says that when we switch the role of attackers and supporters
(replace $g$ with $-g$), the aggregated strength value should just switch sign.
For the special case $\weight=0.5$, the influence condition says 
that a positive aggregate must yield the same distance
to $1$ as the negative aggregate yields to $0$. 
If $\weight\neq 0.5$, there is a natural asymmetry because the initial weight is now
either closer to $0$ or $1$. However, a negative aggregate for weight $w$ should still
yield the same distance to $0$ as the positive aggregate yields to $1$ for weight $1-w$.
In the following proposition, we give a more intuitive interpretation of Duality.
\begin{proposition}
Let $\acceptSemantics$ be a basic modular semantics
that satisfies Duality and 
let $\bag = (\arguments, \weight, \attacks, \supports)$ be a BAG
such that $\acceptSemantics(\bag) = s^* \neq \bot$.
If there are $\arguments_i, \arguments_j$ such that
\begin{enumerate}
	\item $g_i = - g_j$ or, more generally, $\alpha_{g_i}(s^*) = -\alpha_{g_j}(s^*)$,
	\item $\weight_i = 1- \weight_j$,
\end{enumerate}
then  $\acceptSemantics(\bag)_i = 1 - \acceptSemantics(\bag)_j$.
\end{proposition}
\ifnum\ProofVersion=1
\begin{proof}
First note that $g_i = - g_j$ implies $\alpha_{g_i}(s^*) = \alpha_{-g_i}(s^*) = -\alpha_{g_j}(s^*)$
by Duality of the aggregation function.
By the contraction principle, $s^*$ is a fixed-point of $(\iota_{\weight_i} \circ \alpha_{g_i})$
and $(\iota_{\weight_j} \circ \alpha_{g_j})$.
Therefore,
\begin{align*}
s_i^* 
&= \iota_{\weight_i}(\alpha_{g_i}(s^*))
= \iota_{\weight_i}(-\alpha_{g_j}(s^*))
= 1 - \iota_{\weight_j}(\alpha_{g_j}(s^*))\\
&= 1 - s_j^*.
\end{align*}
\end{proof}
\fi
The basic case of the first condition says that $\arguments_i$'s attackers are $\arguments_j$'s supporters
and vice versa. This is intuitive, but somewhat restrictive. The more general version says that the magnitude of the aggregated strength at $\arguments_i$ and $\arguments_j$ is equal, but it acts in
different directions.
The second condition says that the initial weights of $\arguments_i$ and $\arguments_j$ are complementary.
Intuitively, we should then expect that their final strength values will also be complementary.
We illustrate this in the following example.
\begin{example}
Consider the BAG in Figure \ref{fig_duality_bag}.
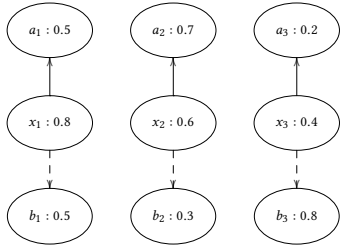
\begin{figure}[tb]
		\begin{center}
		  \scalebox{.6}{
				\xymatrix{
  *+++++[o][F-]{a_1: 0.5} 
	&*+++++[o][F-]{a_2: 0.7}
	&*+++++[o][F-]{a_3: 0.2} \\
  *+++++[o][F-]{x_1: 0.8} \ar@{->}[u]  \ar@{-->}[d]  
	&*+++++[o][F-]{x_2: 0.6} \ar@{->}[u]  \ar@{-->}[d]    
	&*+++++[o][F-]{x_3: 0.4} \ar@{->}[u]  \ar@{-->}[d]   \\
  *+++++[o][F-]{b_1: 0.5}   
	&*+++++[o][F-]{b_2:0.3}   
	&*+++++[o][F-]{b_3:0.8}  
			}	
		}
	\caption{Duality Example. \label{fig_duality_bag}}
	\end{center}
\end{figure}
Table \ref{fig:results_duality_bag} shows the strength values for the three semantics from Table \ref{fig:example_semantics}.
\begin{table}
	\begin{tabular}{lcccccc}
		\hline
		& $a_1$ & $b_1$ & $a_2$ & $b_2$ & $a_3$ & $b_3$  \\[0.0cm]
		Weight $\weight$ & $0.50$ & $0.50$ & $0.70$ & $0.30$ & $0.20$ & 0.80 \\[0.0cm]
		\hline
		Euler & $0.39$ & $0.65$ & $0.63$ & $0.41$ & $0.15$ & $0.84$  \\[0.0cm] 
		DFQ(1) & $0.10$ & $0.90$ & $0.28$ & $0.72$ & $0.12$ & $0.88$  \\[0.0cm]
		QE(1) & $0.30$ & $0.70$ & $0.51$ & $0.49$ & $0.17$ & $0.83$  \\
	\end{tabular}
	\caption{Initial weight and strength values for arguments in Figure \ref{fig_duality_bag} under semantics from Table \ref{fig:example_semantics}.\label{fig:results_duality_bag}}	
\end{table}
The asymmetry of the Euler-based semantics can already be seen from the subgraph with indices $1$.
Whereas the support of $x_1$ increases the strength of $b_1$ by $0.15$, its attack decreases the
strength of $a_1$ only by $0.11$. Both the DF-QuAD algorithm and the quadratic energy model
induce a symmetrical impact for attacks and supports. 

As we move the initial weight away from 
$0.5$, there is a natural asymmetry caused by the fact that the distance from the initial weight to $0$ and $1$ is now different.
However, attack and support should still behave in a dual manner. For the subgraph with indices
$2$, the initial weight of $a_2$ and $b_2$ is moved away from $0.5$ by $0.2$ in different directions.
Again, the increase caused by a support should equal the decrease caused by an attack.
For the DF-QuAD algorithm, the change is $0.42$, for the quadratic energy model $0.19$.
Similarly, for the subgraph with indices $3$, the DF-QuAD algorithm causes a change of $0.08$, 
the quadratic energy model causes a change of $0.03$.
\end{example}

In Table \ref{fig:aggregation_influence_examples}, all building blocks other than the Euler-based
influence functions can be selected in order to satisfy duality as we show in the following proposition. 
\begin{proposition}
The Sum-, Product- and Top-aggregation functions satisfy condition 1 in Definition \ref{def_duality}.
The Linear($\kappa$) and p-Max($\kappa$) influence functions satisfy condition 2 in Definition \ref{def_duality}. 
\end{proposition}
\ifnum\ProofVersion=1
\begin{proof}
Sum: 
\begin{align*}
\alpha^\Sigma_g(s) 
= \sum_{i} g_i \cdot s_i
= - \sum_{i} (-g_i) \cdot s_i 
= - \alpha^\Sigma_{-g}(s).
\end{align*}

Product:
\begin{align*}
\alpha^\Pi_g(s)  
&= \prod_{i: g_i = -1} (1 - s_i) - \prod_{i: g_i = 1}^n (1 - s_i) \\
&= - \big(- \prod_{i: -g_i = 1} (1 - s_i) + \prod_{i: -g_i = -1}^n (1 - s_i)\big) \\
&= -\alpha^\Pi_{-g}(s).
\end{align*}

Top: 
\begin{align*}
\alpha^\textit{max}_g(s) 
&= M_g(s) - M_{-g}(s)
= - (- M_g(s) + M_{-g}(s))\\
&= - \alpha^\textit{max}_{-g}(s).
\end{align*}

Linear($\kappa$):
\begin{align*}
&1 - \iota^l_{(1 - \weight)}(a) \\
&=  \weight + \frac{1 - \weight}{\kappa} \cdot \max \{0, -a\} - \frac{\weight}{\kappa} \cdot \max \{0, a\} \\
&=\iota^l_{\weight}(-a).
\end{align*}

p-Max($\kappa$):
\begin{align*}
&1 - \iota^p_{(1- \weight)}(a) \\
&=  \weight + \frac{1 - \weight}{\kappa} \cdot h(-a) - \frac{\weight}{\kappa} \cdot h(a)\\
&=\iota^p_{\weight}(-a).
\end{align*}
\end{proof} 
\fi
Since the DF-QuAD algorithm and the quadratic energy model are constructed from these building 
blocks, an immediate consequence is that they 
satisfy duality. 

\section{Implementing Modular Semantics with Attractor}

The framework of modular semantics and has been implemented in the Java library Attractor\footnote{\url{https://sourceforge.net/projects/attractorproject}} \cite{potykatutorial}.
The user can initialize modular semantics with different combinations of 
aggregation and influence functions and can use existing implementations of algorithms to compute strength values
using discrete (by using Euler's method with step size $1$) or continuous semantics.
Implementations of the aggregation and influence functions discussed here
already exist, but new functions can be added easily by implementing existing interfaces. 
For example, the semantics of the DF-QuAD algorithm can be initialized with the following three lines of code:
\begin{align*}
&\textit{AggregationFunction agg = new ProductAggregation(); } \\
&\textit{InfluenceFunction inf = new LinearInfluence(1);} \\
&\textit{ContinuousModularModel mod =} \\
& \qquad \textit{new ContinuousModularModel(agg, inf);} 
\end{align*}
Attractor contains implementations of RK4 (for reliable computations) 
and Euler's method (for simulating discrete semantics and illustration purposes). 
Both implementations have a printing variant that automatically generates plots like in
Figure \ref{fig_continuized_convergence_example} (RK4) and
Figure \ref{fig_divergence_example} (Euler) while computing the solution. 
The plots are generated by JFreeChart\footnote{\url{http://www.jfree.org/jfreechart/}}.
For example, in order to use the plotting variant of RK4, we can add the following code:
\begin{align*}
&\textit{AbstractIterativeApproximator approximator = } \\
& \qquad \textit{new PlottingRK4(mod); } \\
&\textit{mod.setApproximator(approximator);} 
\end{align*}
Finally, the strength values for a BAG can be computed. Attractor provides a simple syntax
to define BAGs in text files. 
The file format is inspired by the format used in ConArg\footnote{\url{http://www.dmi.unipg.it/conarg/}} \cite{bistarelli2016conarg}, but adds 
weights and support relations.
BAGs can also be defined programmatically if more flexibility is required. We refer to \cite{potykatutorial} for details
on creating BAGs. Assuming that a BAG file is given, the strength values can be computed by adding the following lines of code:
\begin{align*}
&\textit{BAGFileUtils fileUtils = new BAGFileUtils();} \\
&\textit{BAG bag = fileUtils.readBAGFromFile(file);} \\
&\textit{mod.setBag(bag); } \\
&\textit{mod.approximateSolution(10e-2, 10e-4, true);} 
\end{align*}
The two numerical parameters correspond to the step size and the termination condition, respectively.
Mathematically, the algorithms converge to a fixed-point at which all derivatives will be $0$.
However, even mathematically, the fixed-point may not be reached in finite time.
In practice, we also have to think about numerical accuracy, and so we usually stop when the derivatives are sufficiently small.
Let us emphasize that the user does not have to think about derivatives.
The derivatives are given by the differential equations. When adding new aggregation or 
influence functions, the differential equations are automatically derived as explained
in Proposition \ref{prop_continuizing_independent_semantics}.  
The logic is already implemented in the class \emph{ContinuousModularModel}.
So when implementing a new aggregation or influence function, only the logic for
aggregating strength values or adapting the initial weight needs to be implemented.

\section{Related Work}

In the original abstract argumentation framework \cite{dung1995acceptability}, arguments can only be attacked by other arguments.
\emph{Bipolar argumentation frameworks} \cite{amgoud2004bipolarity,OrenN08,cayrol2013bipolarity} add a
support relation.
Classical semantics can only accept or reject arguments \cite{baroni2011introduction},
but various proposals have been made to allow for a more fine-grained evaluation.
Among others, it has been suggested to apply tools from probabilistic reasoning
\cite{dung2010towards,li2011probabilistic,rienstra2012towards,hunter2014,doder2014probabilistic,polberg2014probabilistic,hunter2014probabilistic,Prakken18,HunterPolbergPotyka18,RienstraEtAl18}
or to rank arguments based on fixed-point equations \cite{besnard2001logic,leite2011social,correia2014efficient,barringer2012temporal} or the graph structure \cite{cayrol2005graduality,amgoud2013ranking}.

In recent years, several weighted bipolar argumentation frameworks as considered here have been presented \cite{baroni2015automatic,rago2016discontinuity,amgoud2017evaluation,mossakowski2018modular,potyka2018Kr}. 
The QuAD algorithm from \cite{baroni2015automatic} was designed to evaluate the strength
of answers in decision-support systems. However, it can show discontinuous behaviour
that is undesirable in some cases. The DF-QuAD algorithm (Discontinuity-free QuAD) \cite{rago2016discontinuity} was proposed as an alternative that avoids this behaviour. 
Some additional interesting semantical guarantees are given by the Euler-based semantics that was introduced in \cite{amgoud2017evaluation}.
The QuAD algorithms mainly lack these properties due to the fact that their aggregated strength values \emph{saturate}.
That is, as soon, as an attacker (supporter) with strength $1$ exists, the other attackers (supporters) become irrelevant for the aggregated value.
The Euler-based semantics avoids many problems, but has some other drawbacks that can be undesirable. 
Arguments initialized with strength $0$ or $1$ remain necessarily unchanged
under Euler-based semantics and, as we saw, attacks and supports have an asymmetrical impact.
The \emph{quadratic energy model} introduced in \cite{potyka2018Kr} avoids these problems.
In \cite{mossakowski2018modular}, some other related models have been studied that use initial weights,
an aggregation and an influence function as well, but the final strength values can also take values 
from the interval $[-1,1]$ or general real numbers. Other aggregation and influence functions
for these cases have been discussed in \cite{mossakowski2018modular} as well.

A first collection of general axioms for weighted bipolar frameworks has been presented in \cite{amgoud2017evaluation}.
Several authors noted recently that the axioms can be simplified by using more elementary properties \cite{mossakowski2018modular,baroni2018many,amgoud2018gradual}.
The idea of modular semantics from \cite{mossakowski2018modular} seems particularly useful
because it allows creating new semantics with interesting guarantees by simply combining suitable aggregation and influence functions. This approach bears some resemblance to representation theorems
considered in other fields that relate semantical properties of operators to elementary properties
of functions that can be used to create these operators. 
Some ideas similar to modular semantics have been invented independently for the special case where only attack relations are present in \cite{amgoud2018gradual}.

\section{Discussion and Future Work}

We extended the framework of modular semantics from \cite{mossakowski2018modular}
in several directions. Our main focus was on convergence guarantees. We generalized the convergence
guarantees from \cite{mossakowski2018modular} to Lipschitz-continuous aggregation and influence functions.
This allowed us, in particular, to derive convergence guarantees for semantics based on 
product-aggregation like the DF-QuAD algorithm.
We also complemented the results from \cite{mossakowski2018modular} with runtime guarantees based on the approximation accuracy and the Lipschitz constants. 
The Lipschitz constants provided in Table \ref{fig:aggregation_influence_examples} can be used to derive
further convergence guarantees in combination with Proposition \ref{prop_convergence_criterion}.
There are many other interesting candidates for aggregation and influence functions
and, provided that they are Lipschitz-continuous, Proposition \ref{prop_convergence_criterion} can be applied
to derive convergence guarantees easily.
For example, truncated sums like the Lukasiewicz T-conorm could be interesting.
In combination with the linear influence function they can guarantee that the extreme values $0$ and
$1$ are taken in desirable cases (e.g., if there is only one attacker/supporter with strength $1$) while avoiding the saturation property of the QUAD algorithms. 

As we discussed, convergence guarantees for discrete models are often bought at the expense of 
open-mindedness. 
We demonstrated that 
we can avoid divergence problems without giving up open-mindedness
by continuizing discrete models as proposed in \cite{potyka2018Kr}. 
It is currently an open question if and under which conditions 
continuous models converge for general cyclic BAGs,
but until now, no divergence examples have been found.
The
continuization of all basic modular semantics yields a well-defined continuous model
as Proposition \ref{prop_continuizing_independent_semantics} explains.
The limits of discrete and continuized models are guaranteed to be equal for acyclic
BAGs and for cyclic BAGs that induce a contractive update function. 
Further investigations are necessary, but it currently seems 
that whenever a discrete model converges, the continuized model converges to the same solution. 

Semantically, we complemented modular semantics with the Duality property.
After relating this property to elementary properties of aggregation and influence functions, 
it can be checked more easily. We showed, in particular, that it is satisfied by DF-QuAD.

Finally, we explained how weighted argumentation problems can be solved with the Java library Attractor. 
Modular semantics allow for very convenient abstractions. Dependent on the user's expertise, new semantics can be implemented completely from scratch, 
can be constructed from self-implemented aggregation and influence functions 
or by just combining pre-implemented aggregation and influence functions.
A graphical user interface is work in progress. 

\bibliographystyle{ACM-Reference-Format}
\bibliography{references}

\end{document}